\documentclass[final]{cvpr}

\usepackage{times}
\usepackage{epsfig}
\usepackage{graphicx}
\usepackage{amsmath}
\usepackage{amssymb}
\usepackage{appendix}

\usepackage[pagebackref=true,breaklinks=true,colorlinks,bookmarks=false]{hyperref}
\usepackage{multirow}
\usepackage{subcaption}
\usepackage{pifont}
\usepackage{paralist}
\usepackage{float}

\def\cvprPaperID{2208} 
\def\confYear{CVPR 2021}

\makeatletter
\newcommand{\printfnsymbol}[1]{%
  \textsuperscript{\@fnsymbol{#1}}%
}
\makeatother

\begin{document}

\title{GAIA: A Transfer Learning System of Object Detection that Fits Your Needs}



\author{
Xingyuan Bu $^{2,4}$\thanks{Equal contributions.}
\and
Junran Peng $^{1,3}$\printfnsymbol{1}
\and
Junjie Yan $^{4}$
\and
Tieniu Tan $^{1,3}$
\and
Zhaoxiang Zhang $^{1,3}$\thanks{Corresponding author.}
\and
$^1$University of Chinese Academy of Sciences, $^2$Beijing Institute of Technology
 \\
$^3$Center for Research on Intelligent Perception and Computing, CASIA, $^4$SenseTime Group Limited
\and
\texttt{xingyuanbu@gmail.com}, \texttt{jrpeng4ever@126.com}\\
\texttt{yanjunjie@sensetime.com},
\texttt{zhaoxiang.zhang@ia.ac.cn}, \texttt{tnt@nlpr.ia.ac.cn}
}

\maketitle

\begin{abstract}
Transfer learning with pre-training on large-scale datasets has played an increasingly significant role in computer vision and natural language processing recently. 
However, as there exist numerous application scenarios that have distinctive demands such as certain latency constraints and specialized data distributions, it is prohibitively expensive to take advantage of large-scale pre-training for per-task requirements.
In this paper, we focus on the area of object detection and present a transfer learning system named GAIA, which could automatically and efficiently give birth to customized solutions according to heterogeneous downstream needs.
GAIA is capable of providing powerful pre-trained weights, selecting models that conform to downstream demands such as latency constraints and specified data domains, and collecting relevant data for practitioners who have very few datapoints for their tasks.
With GAIA, we achieve promising results on COCO, Objects365, Open Images, Caltech, CityPersons, and UODB which is a collection of datasets including KITTI, VOC, WiderFace, DOTA, Clipart, Comic, and more.
Taking COCO as an example, GAIA is able to efficiently produce models covering a wide range of latency from 16ms to 53ms, and yields AP from 38.2 to 46.5 without whistles and bells.  
To benefit every practitioner in the community of object detection, 
GAIA is
released at \href{https://github.com/GAIA-vision}{https://github.com/GAIA-vision}.
\end{abstract}

\section{Introduction}

\begin{figure}[ht]
\centering
\begin{subfigure}[b]{0.42\textwidth} 
\centering
\includegraphics[width=\textwidth]{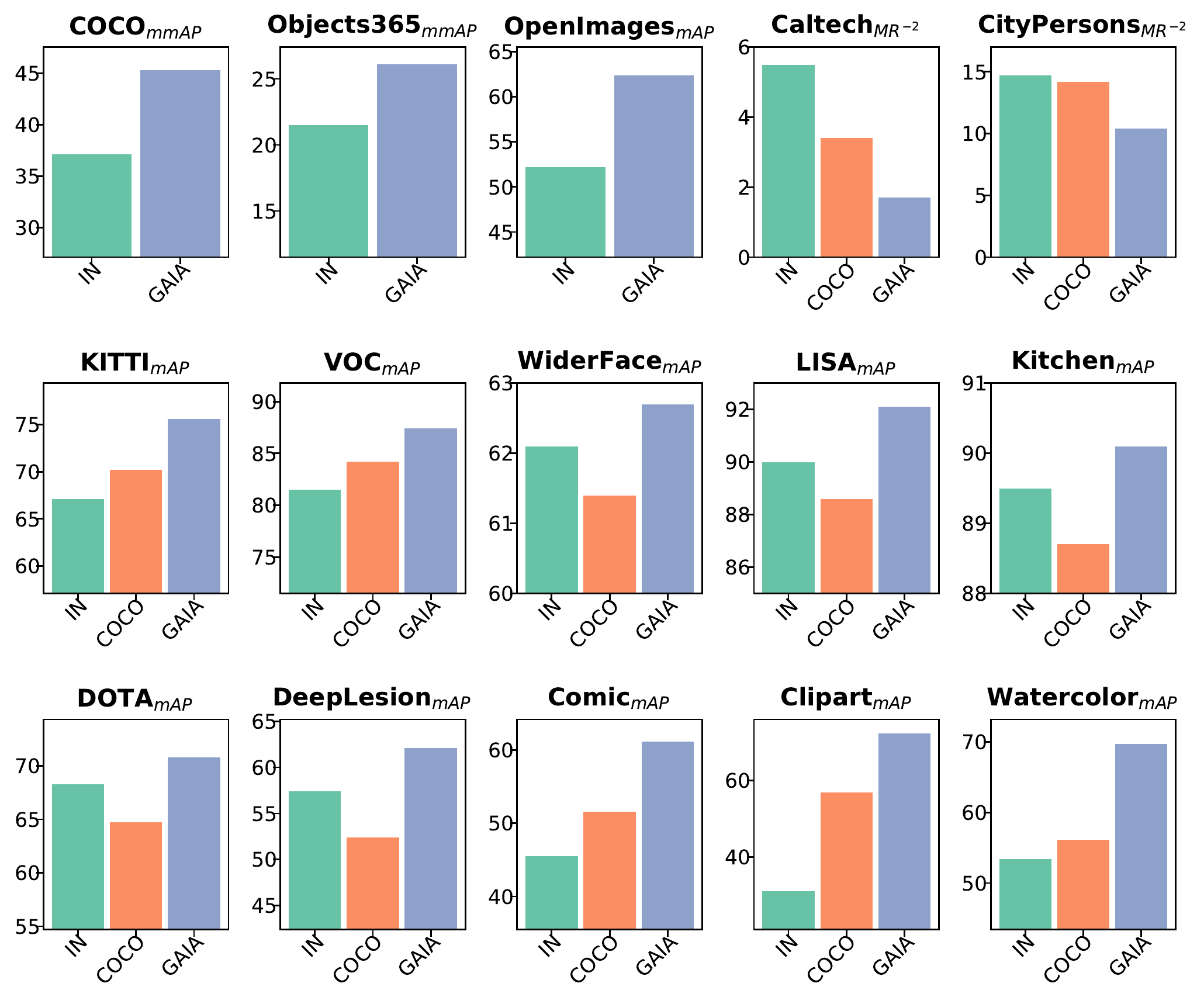}
\vspace{-0.5cm} 
\caption{
Comparison of models produced by GAIA, and ResNet50 pre-trained on ImageNet and COCO to various downstream tasks. All models share the same latency. For $MR^{-2}$ in Caltech and CityPersons, lower is better.
}
\label{fig:intro:main_results_15in1}
\end{subfigure}
\centering
\begin{subfigure}[b]{0.43\textwidth} 
\centering
\includegraphics[width=0.98\textwidth]{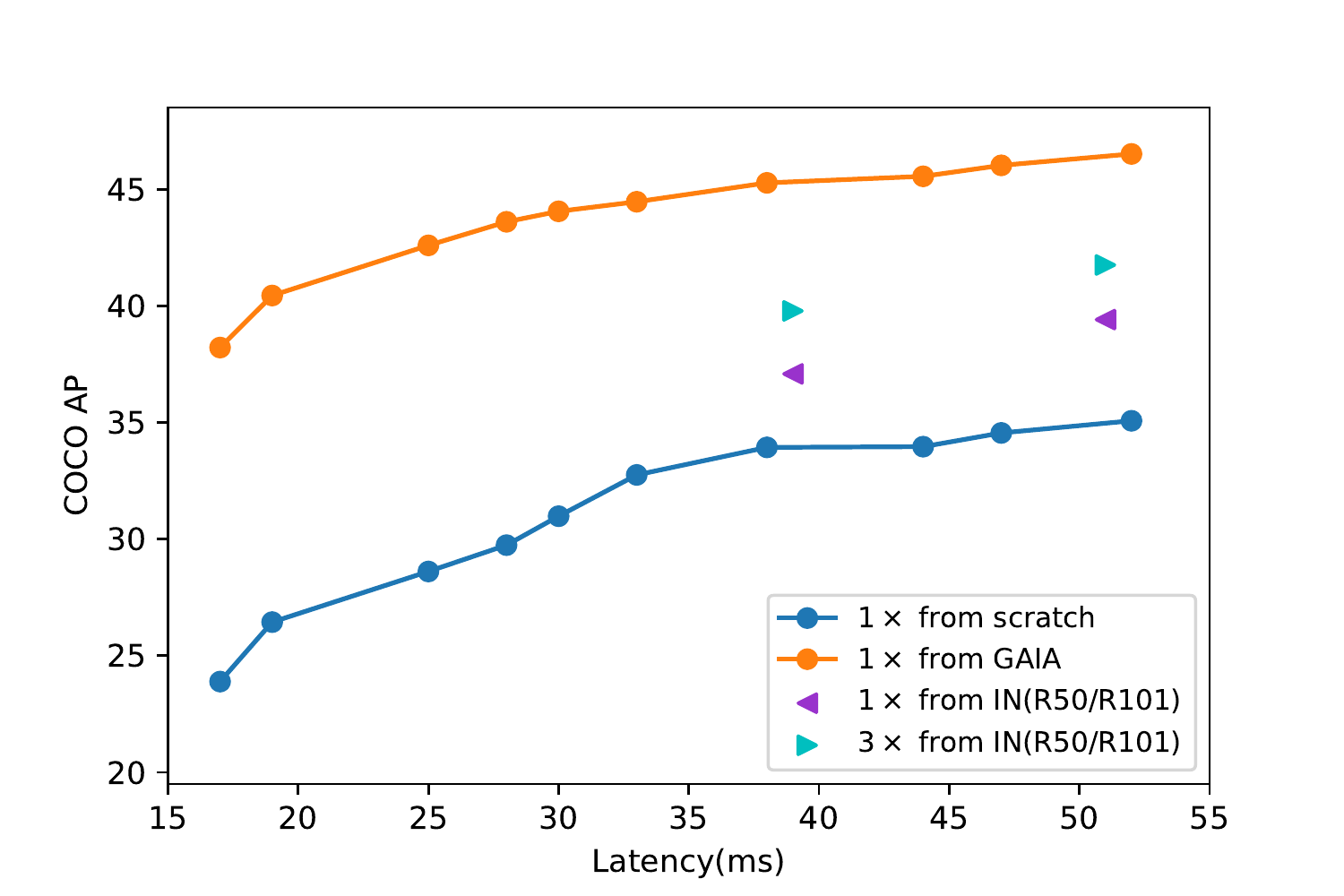}
\caption{Comparison of models with different latency on COCO.}
\label{fig:intro:main_results_latency}
\end{subfigure}
\vspace{-0.2cm} 
\caption{
Transfer performance of GAIA models that adapt to various downstream needs including specified domains and latency constraints. No whistles and bells are used. 
}
\label{fig:intro:main_results}
\vspace{-0.3cm} 
\end{figure}

Transfer learning has been one of the most powerful techniques throughout the history of deep learning. The paradigm of pre-training on large-scale source datasets and finetuning on target datasets or tasks is applied over a wide range of tasks in both computer vision (CV) and natural language processing (NLP). 
The motivation behind is to endow models the general-purpose abilities and transfer the knowledge to downstream tasks, in order to acquire higher performance or faster convergence. 
Until recently, the influence of transfer learning has moved onto the next level as the scales of data are growing exponentially. 
In BiT~\cite{Kolesnikov2020BigT} and WSL~\cite{Mahajan2018ExploringTL}, pre-training on enormous data like JFT-300M and Instagram (3.5B images) has been proved to yield extraordinary improvements over the conventional ImageNet pre-training on downstream tasks. Similar trends are going on in NLP, as shown in BERT~\cite{Devlin2019BERTPO}, T5~\cite{Raffel2019ExploringTL} and GPT-3~\cite{Brown2020LanguageMA}.

Although transfer learning with huge scale of pre-training data has shown the great success, it is severely afflicted by its inflexibility on model architecture. As stated in the ``no free lunch'' theorem~\cite{Wolpert1997NoFL}, no single algorithm is best suited for all possible scenarios and datasets. Different datasets may request for different model architectures, and different application scenarios may request for model of different scales. To take advantage of the transfer learning, these customized models are obliged to be trained from scratch on the whole upstream datasets, which is prohibitively expensive.

This issue is even more serious in object detection, as it is one of the most significant computer vision tasks and covers an wide range of deployment scenarios. 
In practice, detectors are always supposed to work across various devices with distinctive resource constraints, which requires detectors to have corresponding input sizes and architectures.
In addition, the correlation between distributions of object scales and the adapted network architectures are very close in object detection.
Therefore, the demand for task-specific architecture adaptation is much stronger in object detection than other tasks such as image classification or semantic segmentation.

In this paper, we introduce a transfer learning system in object detection that aims to harmonize the gap between large-scale upstream training and downstream customization.
We name this system ``GAIA'' as it could swiftly give birth to a variety of specialized solutions to heterogeneous demands, including task-specific architectures and well-prepared pre-trained weights. For users who have very few datapoints for their tasks, GAIA is able to collect relevant data to fuel their needs.
Our objective is not to propose a specific method, but more to present an integrated system that brings convenience to practitioners in the community of object detection.
There are two components in GAIA: {\it task-agnostic unification} and {\it task-specific adaptation}.

To begin with, we conduct the task-agnostic unification on data and architectures respectively. In order to unleash the potential of transfer learning, we collect data from multiple sources and build a huge data pool with a unified label space. The unified label space is formulated based on the {\it word2vec} similarity, which prevents knowledge conflicts among duplicated categories from distinctive sources and enables data of relevant categories to jointly boost the performance of detector. Besides, covering a wide range of categories provides indicators for task-specific adaptation. 
To realize the purpose of training plenty of models efficiently on huge upstream data, we adopt the weight sharing scheme which has been widely used in~\cite{Liu2019DARTSDA, Bender2018UnderstandingAS, Pham2018EfficientNA, Huang2016DeepNW, Yu2019UniversallySN, cai2020once, Yu2020BigNASSU}, which enables models of different widths and depths to be optimized together. As models may interfere with each other during collective training, we propose an ``anchor-based progressive shrinking'' to alleviate the problem.

In the task-specific adaptation procedure, GAIA needs to find adapted model architectures according to the given tasks. We quantitatively assess the ranking ability of different search methods based on the Kendall Tau measure~\cite{Kendall1938ANM} and propose an efficient and reliable method of selecting models that surprisingly fit the downstream task, regardless of data domains or latency constraints (Figure~\ref{fig:intro:main_results}).    
To extend the utility of GAIA on the ubiquitous data-scarce scenarios, we develop GAIA an ability of collecting relevant data to downstream tasks from data pool, which yields further improvements.

The contributions of our paper are as follows:
\begin{itemize}
    \item We demonstrate how transfer learning and weight sharing learning could be well combined, to produce powerful pre-trained models over a variety of architectures simultaneously.
    \item We propose an efficient and reliable approach of finding the adapted network architectures to specified downstream tasks. Powered by pre-training and task-specific architecture selection, GAIA achieves surprisingly good results over 10 downstream tasks without exclusive tuning on hyper-parameters.
    \item GAIA has the capability of finding relevant data based on 2 images per category in the downstream tasks to support finetuning. This further extends the utility of GAIA in data-scarce settings.
\end{itemize}

\section{Related Work}

\subsection{Object Detection}
Beginning with R-CNN~\cite{Girshick2014RichFH} and its predecessors like Fast-RCNN~\cite{Girshick2015FastR} and Faster-RCNN~\cite{Ren2015FasterRT}, deep learning grows prosperously in the area of object detection. A great amount of methods are proposed to advance the area, including Mask RCNN~\cite{He2020MaskR}, FPN~\cite{Lin2017FeaturePN}, DCN~\cite{Dai2017DeformableCN} and Cascade RCNN~\cite{Cai2018CascadeRD}. 
Since it has been years that the most researches are conducted on Pascal VOC or MS-COCO datasets, which are viewed as ``small data" by modern standards, there are researchers begin to explore on more challenging issues such as cross-domains object detection or large-scale object detection in the wild.
Wang \etal~\cite{Wang2019TowardsUO} develop an universal object detection system that works across a wide range of domains including traffic signs to medical CT images. With the advent of larger datasets of object detection such as Objects365~\cite{shao2019objects365}, Open Images~\cite{Kuznetsova2020TheOI}, and Robust Vision~\cite{robustvision}, there are works~\cite{Peng2020LargeScaleOD} focusing on solving issues in this scenarios such as long-tailed data distribution and multi-labels problems.

\subsection{Neural Architecture Search}
Neural architecture search(NAS) aims at automating the architecture of network design process under certain constraints.
Earlier methods~\cite{Zoph2017NeuralAS, Zoph2018LearningTA,baker2017designing,cai2018efficient} train thousands of candidate networks with distinctive architectures and rely on reinforcement learning or evolution algorithm to discover the optimal architectures. These methods mostly require unimaginably huge computation resources and seem forbidding for most research institutions.
Gradient-based methods like DARTS~\cite{Liu2019DARTSDA} and Proxyless~\cite{cai2018proxylessnas} come out that greatly alleviate the problem through training and searching candidate architectures inside a single super-net. However, in real world that heterogeneous tasks exist, these methods requires repetitive searching and training process for each task according to the specified hard-ware platforms and latency constraints.
To address this issue, researchers further propose methods~\cite{Yu2019UniversallySN, cai2020once, Yu2020BigNASSU} that are to produce models across different inference latency all at once.  
As in area of object detection, NAS methods also spring up such as ~\cite{Chen2019DetNASBS, Peng2019EfficientNA, Ghiasi2019NASFPNLS, Tan2020EfficientDetSA}.

\subsection{Transfer Learning}
Transfer learning has been playing a non-negligible role throughout the history of deep learning. 
The paradigm of pre-training on ImageNet~\cite{Deng2009ImageNetAL} dataset has immensely pushed forward the development of computer vision, covering a wide range of tasks such as object detection~\cite{Girshick2014RichFH,Girshick2015FastR,Ren2015FasterRT,Liu2016SSDSS,Lin2017FocalLF,Redmon2016YouOL},semantic segmentation~\cite{Long2015FullyCN, Chen2018DeepLabSI}, pose estimation~\cite{Toshev2014DeepPoseHP} and \etc. 
Recently, there are studies~\cite{Sun2017RevisitingUE, Mahajan2018ExploringTL, Yalniz2019BillionscaleSL, Kolesnikov2020BigT} focusing on transfer learning on larger scale of data such as JFT-300M and Instagram (3.5B images). With the help of weakly supervised learning~\cite{Mahajan2018ExploringTL}, knowledge distillation~\cite{Yalniz2019BillionscaleSL} and supervised learning~\cite{Kolesnikov2020BigT}, surprisingly good results are achieved.
As in object detection, ~\cite{shao2019objects365, Li2019AnAO} also prove the effectiveness of large-scale data.
In addition to transfer pre-trained models, there are methods transferring data directly. In~\cite{Gururangan2020DontSP, Aharoni2020UnsupervisedDC}, the authors show that more in-domain data could benefit diverse NLP tasks. In~\cite{Cui2018LargeSF, yan2020neural, Hasan2020PedestrianDT}, extra similar data is selected to provide better pre-trained models. Domain transfer by adversarial training~\cite{Inoue2018CrossDomainWO} also mitigates the lack of target data.

\section{GAIA}
We introduce GAIA, a transfer learning framework, and its detailed implementation in this section. GAIA consists of two major components: {\it task-agnostic unification} and {\it task-specific adaptation}. 
In the task-agnostic unification part, we collect data from multiple sources and build a large data pool with a unified label space. 
Then we utilize the technique of weight sharing learning for training a supernet, which enables models of various architectures to be optimized collectively. 
In the task-specific adaptation part, GAIA search a most adapted architecture for the given downstream task, initialize the network with weights extracted from the pre-trained supernet, and finetune it on the downstream data. We call this process ``task-specific architecture selection'' (TSAS). 
In addition, to help with users who have very few datapoints for their tasks, GAIA is able to collect the most correlated data to the given tasks from the huge data pool as relevant data. We call this ability of GAIA as ``task-specific data selection'' (TSDS).

\subsection{Task-agnostic Unification}

\subsubsection{Unified Data and Label Space}

Although steady progresses have been made in distinctive datasets, they are independent of each other and often restricted to specific domains~\cite{robustvision}. Therefore, to push forward real-world usability of an object detection system and reduce dataset bias, we merge multiple datasets into a huge data pool with a unified label space $\cup{L}$.
We start with $N$ datasets $D=\{d_1, d_2, \cdots, d_N\}$ and their label spaces $L=\{l_1, l_2, \cdots, l_N\}$. Each label space $l_i$ consists of the categories $\{c_{i1}, c_{i2}, \cdots, c_{i\left|l_i\right|}\}$ corresponding to dataset $d_i$.
To obtain the unified label space, we first choose the largest label space among $L$ as the initial $\cup{L}=\{c_{\cup 1}, c_{\cup 2}, \cdots\}$.
Then we map other label spaces into $\cup{L}$. We mark the $p$-th category $c_{ip}$ from dataset $d_i$ as an identical category to $c_{\cup{q}}$ if their {\it word2vec}~\cite{rehurek_lrec} similarity is higher than a threshold of 0.8. If there is no similarity greater than 0.8 for $c_{ip}$, we mark it as a novel category and append $c_{ip}$ into $\cup{L}$.
Finally, all candidate mappings would be verified for reliability.

In the follow-up development, the unified label space is unique but not static as GAIA has the will to cover as many datasets as possible. When an unseen dataset needs to be merged, we repeat the label space mapping process as mentioned above. To fit the pre-trained network to the updated label space, the last {\it fc} layer extends corresponding new ways while the others remain unchanged.

Using this unified label space rather than training with multiple heads or separate label definitions, reduces the cost of dataset combination and mitigates the potential conflicts between duplicate categories. More importantly, it enables GAIA to provide the indicators for diverse downstream tasks; otherwise, it is impractical to obtain a reliable measure for novel datasets.
The unified label space may introduce long-tail and partial annotation problems as claimed in~\cite{zhao2020object}. Nevertheless, we note that these problems barely influence the downstream finetuning in our experiments.

\begin{figure*}[t]
\centering
\begin{subfigure}[b]{0.3\linewidth} 
\centering
\includegraphics[width=\textwidth]{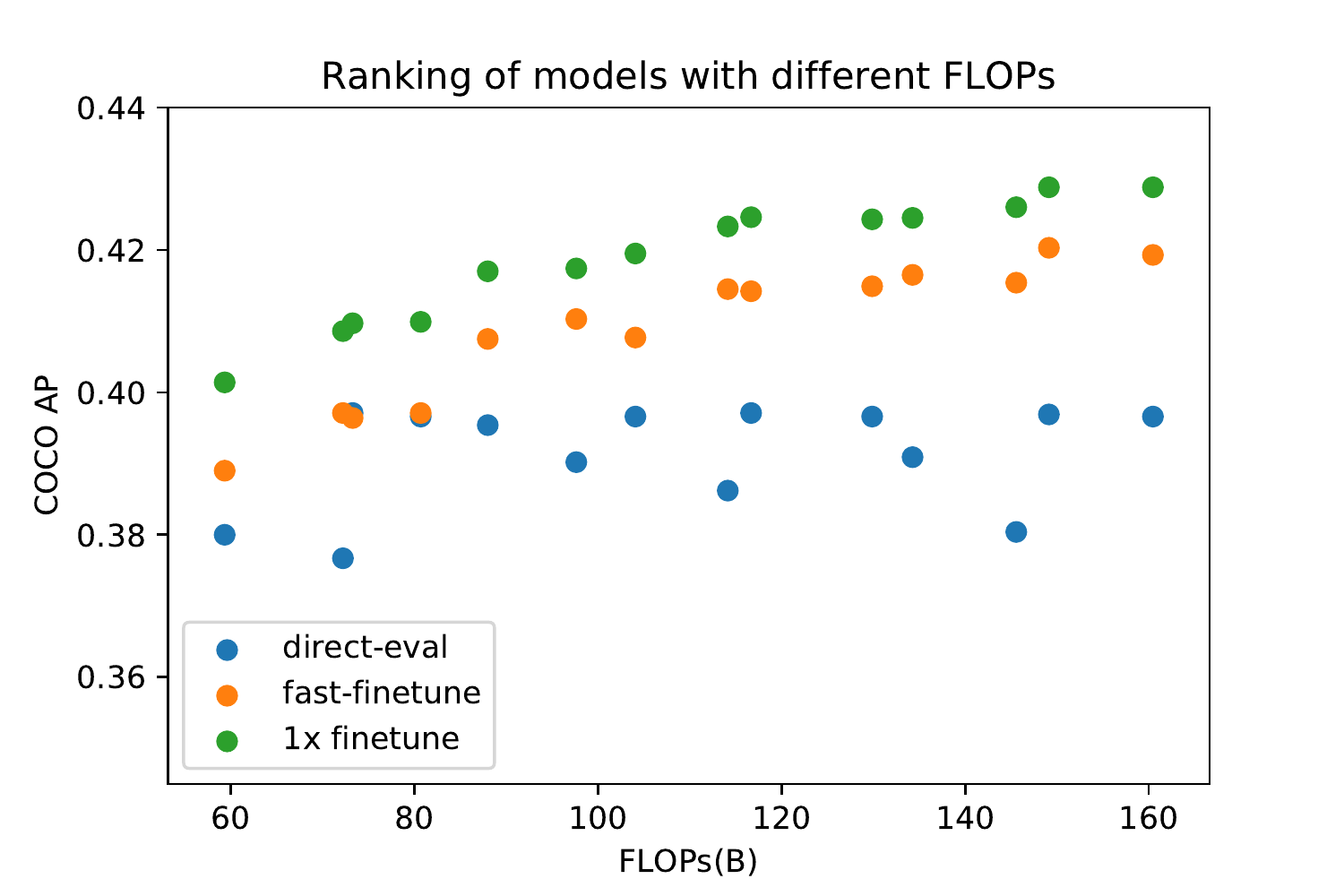}
\caption{}
\label{fig:methods:ranking_different_flops}
\end{subfigure}
\begin{subfigure}[b]{0.3\linewidth} 
\centering
\includegraphics[width=\textwidth]{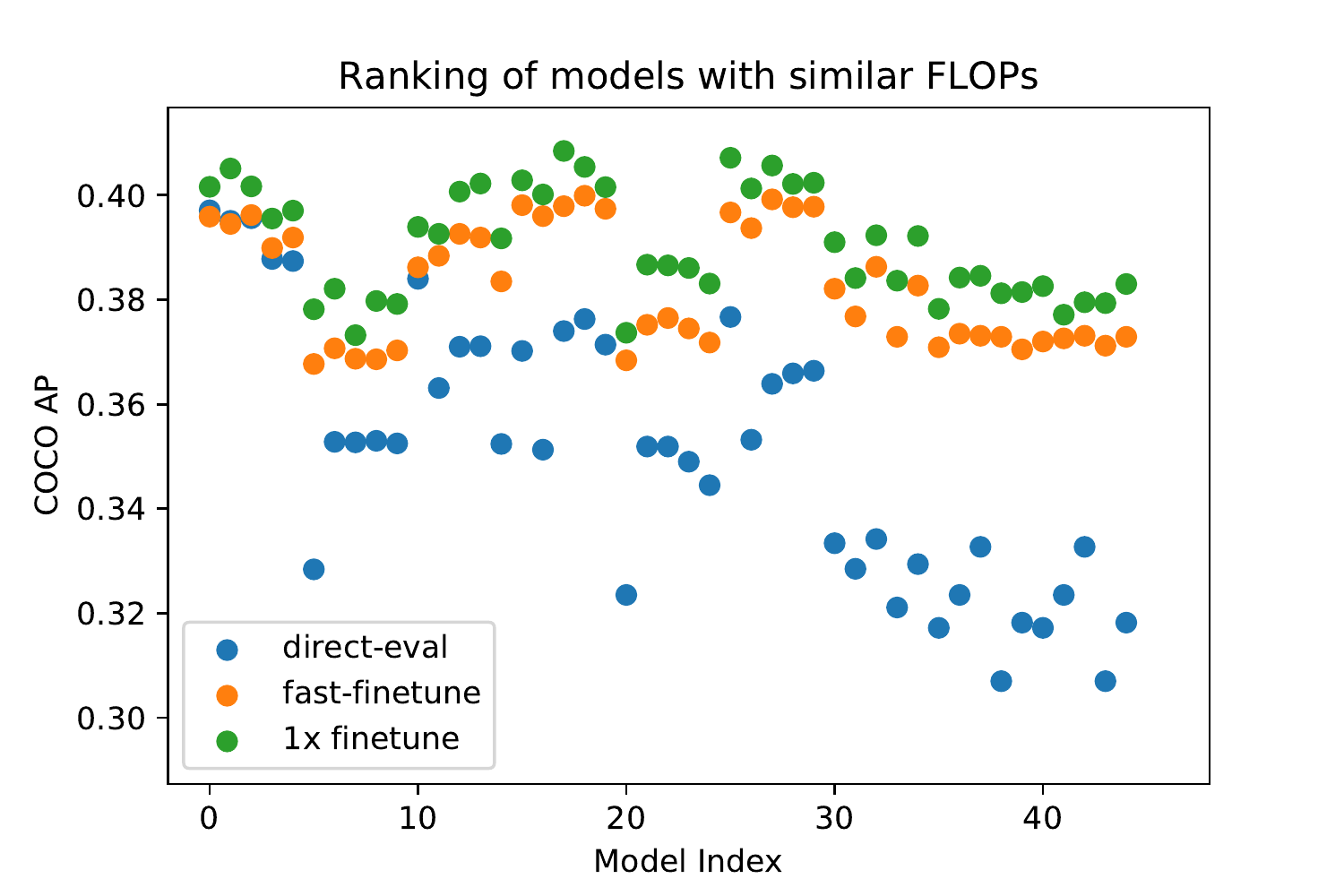}
\caption{}
\label{fig:methods:ranking_similar_flops}
\end{subfigure}
\begin{subfigure}[b]{0.3\linewidth} 
\centering
\includegraphics[width=\linewidth]{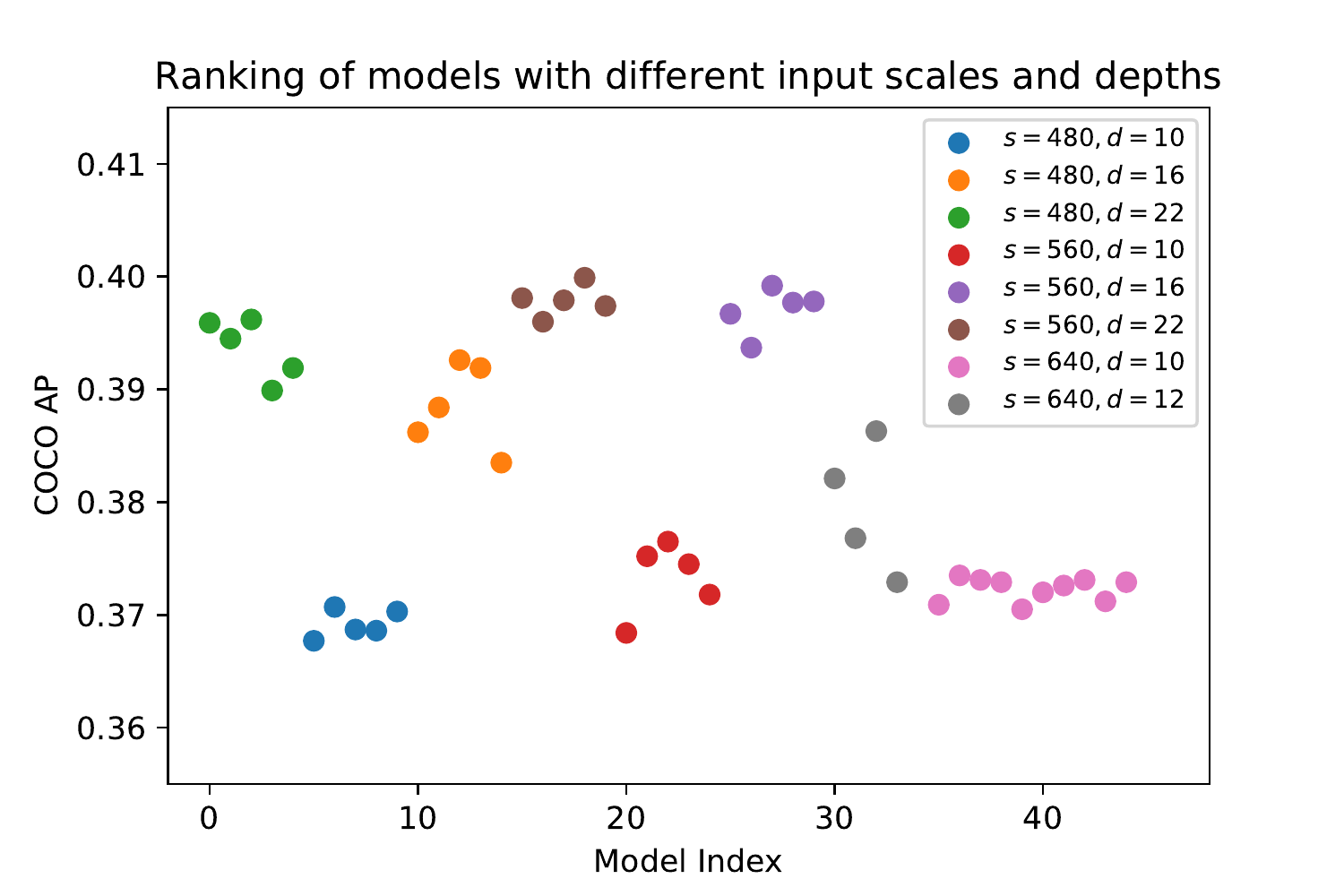}
\caption{}
\label{fig:methods:ranking_different_scale_depth}
\end{subfigure}
\vspace{-0.1in} 
\caption{
Model ranking in different settings. (a) and (b) show the ranking of models with different and similar FLOPs respectively. (c) shows the ranking of ``fast-finetuned'' models with similar FLOPs but different input scales and depths. 
}
\label{fig:methods:model_rankings}
\vspace{-0.1in} 
\end{figure*}

\subsubsection{Unified Architecture Training}
The search space of supernet in GAIA consists of network depth, layer width, and input scale, which are factors mostly correlated with computation budgets. We follow the regime of weight sharing learning mentioned in~\cite{Yu2020BigNASSU}, that only layers and channels of the lower index are kept when sampling subnets out of supernet. 
In this way, models of different architectures could be optimized collectively.
However, there exists a disturbing problem about  weight sharing learning, that different sub-networks are interfering with each other throughout the training process of supernet. 

To mitigate the problem, a method named {\it progressive shrinking} (PS) is proposed in OFA~\cite{cai2020once}, that the search space is gradually loosed, from kernel size to depth, and finally to width. Although it has shown significant effectiveness in task of image classification, it seems not to fit well in object detection. 
The reason underlying has been mentioned in previous works~\cite{Liu2018ReceptiveFB,  Dai2017DeformableCN} that the effective receptive field of network matters in object detection, thus a reasonable search space is required to cover a wide range of network depth. The regime that shrinks search space in order of different aspects might not be able to alleviate the interference among models with huge variation in depth, as in this paper the depth of a single stage may vary from 12 to 87. In addition, it is crucial to keep a good compatibility between input scale and depth as mentioned in~\cite{Tan2020EfficientDetSA} when sampling models during upstream training. 

Taking these understandings into consideration, we propose a training scheme named {\it anchor-based progressive shrinking} (ABPS) that shrinks multiple search dimensions step by step.
To begin with, we select a model anchor and build a subordinate search space surrounding it while keeping the basic compatibility of depth and input scales. After training for a while, we shrink the model anchor, then finetune on the new search space surrounding the model anchor. We repeat the process several times until the entire search space covers a wide latency range.

\subsection{Task-specific Adaptation}

\subsubsection{Task-specific Architecture Selection (TSAS)}
After training supernet on the unified data pool, one needs to select high-quality models based on domains of interests and constraints on computation budgets. 
There two reasons making the architecture selection difficult: First, evaluating more than 300K candidates on each downstream task is prohibitively expensive. Besides, directly evaluate models sampled from supernet may not reflect the true quality of architecture , which is an avoided issue in methods~\cite{Yu2020BigNASSU, cai2020once}. In this part, we delve into the model ranking correlation and seek for a reliable selection regime that exhibits strong ranking ability. The Kendall Tau~\cite{Kendall1938ANM} index is applied to analyze the ranking correlations quantitatively.

To begin with, we focus on how the ranking correlation behaves among models with different FLOPs, as in most cases, models with larger FLOPs tend to have better precisions if they share similar architecture. 
Given a search space surrounding a randomly picked model anchor, we split subnets into groups based on BFLOPs, and study the model ranking across different FLOPs.
In each group, we randomly evaluate 100 models and pick the best-performing models as the representative. 
We finetune these models with the standard $1\times$ scheduler and take the results as ranking references. Then we apply a $0.2\times$ ``fast-finetuning'' scheduler on them and compare the ranking ability with direct evaluation. 
As shown in Figure~\ref{fig:methods:ranking_different_flops}, the correlations between results of direct evaluation and $1\times$ are extremely weak and the Kendall Tau is only 0.18, while  fast-finetuning dramatically strengthen the correlations and the Kendall Tau is boosted from $0.18$ to $0.85$. 
Similar phenomenon is observed when models share the similar FLOPs, as shown in Figure~\ref{fig:methods:ranking_similar_flops}.
Thus in GAIA, we rely on results of the fast-finetuned models as indicators for model selection instead of the direct evaluation results, which are totally unreliable.

Since applying fast-finetuning on 300K models is still costly, we need to narrow down the search space to a tiny but instructive one. 
To this end, we randomly sample models of different input scales and depths around a fixed FLOPs, and apply the fast-finetuning to find the decisive factors. As illustrated in Figure~\ref{fig:methods:ranking_different_scale_depth}, models with the same input scales and depths tend to attain similar precisions while models with either different input scales or different depths have more diverse precisions. 
Thus in GAIA, we apply a two-step search scheme to robustly find high-quality models. In the first step, we randomly sample $K$ (usually 5) models as a group for each combination of input scales and total depths in each sub search space while keep these models satisfying the given constraints.
We directly evaluate these models and pick the best-performing model in each group.
In the second step, we keep the top $50\%$ models among the picked, fast-finetune them and select the best architecture based on the fast-finetuning results.

\subsubsection{Task-specific Data Selection (TSDS)}
\begin{figure}[t]
\begin{center}
\includegraphics[width=0.90\linewidth]{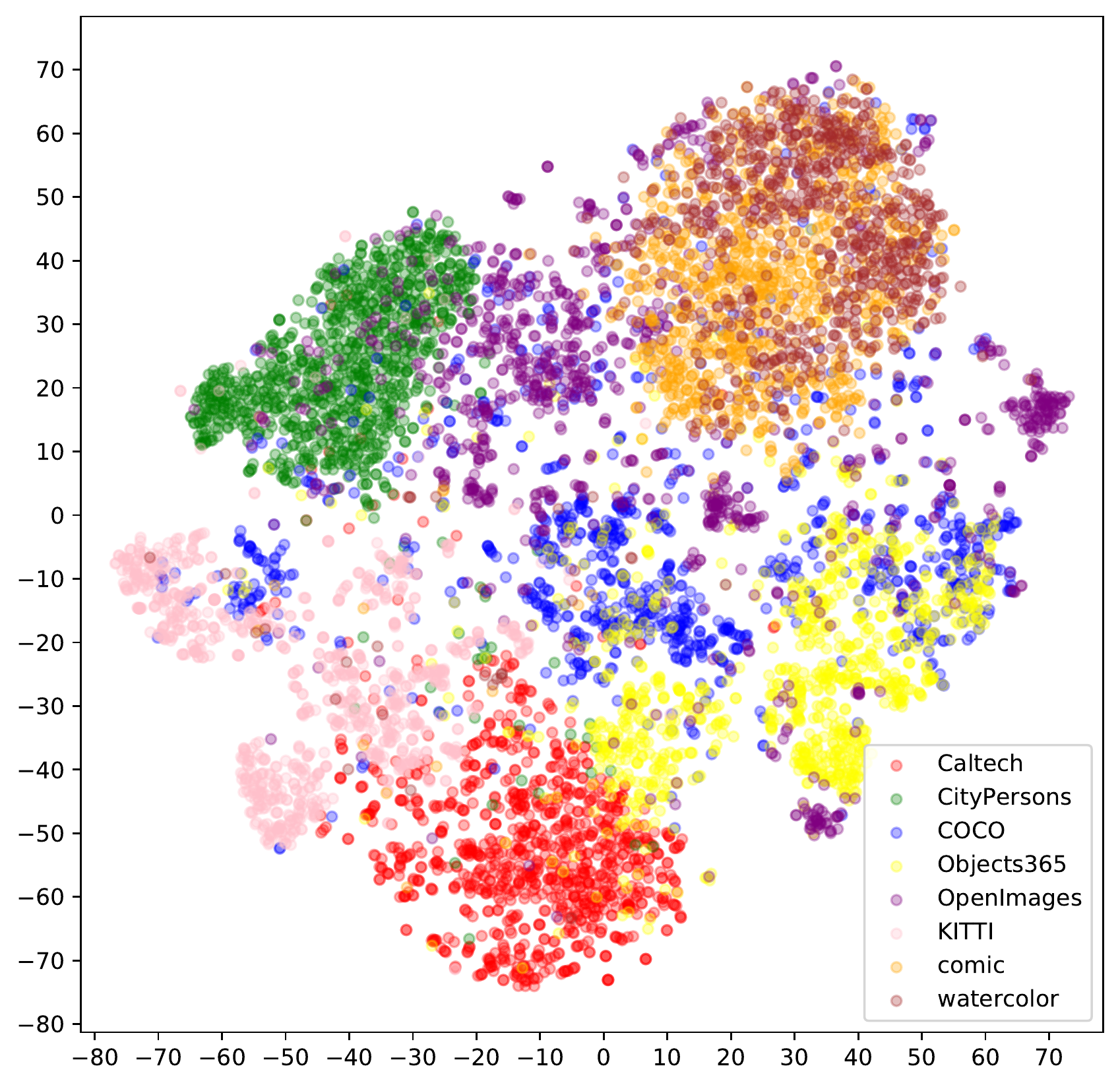}
\end{center}
\vspace{-0.25in} 
\caption{Visualization of output features in {\it fc7} layer from GAIA using t-SNE. We take {\it person} category as an example and sample 1000 images for each dataset. The color represents the dataset for each image. 
}
\label{fig:gaia_datasets_cluster}
\vspace{-0.2in} 
\end{figure}

Since GAIA has been able to produce a series of superior architectures according to downstream tasks, 
we naturally want to take a step further: Is it possible that GAIA could select relevant data to support downstream tasks even more? 
This is particularly favorable in data-scarce scenarios where very few datapoints are available for the tasks.  
Given massive upstream datasets $D_{s}$ and a specific downstream dataset $D_{t}$, data selection aims to find the subset $D_{s}^{*} \in \mathcal{P}\left(D_{s}\right)$, where $\mathcal{P}\left(D_{s}\right)$ is the power set of $D_{s}$, such that $D_{s}^{*}$ minimizes the risk of a model $\mathcal{F}$ on the task dataset $D_{t}$:
\begin{equation}
\label{eq:data_selection_definition}
\begin{aligned}
D_{s}^{*} &= \mathop{\arg\min}_{D_{s}^{*} \in \mathcal{P}\left(D_{s}\right)} \mathbb{E}_{D_{t}}\left[\mathcal{F}\left(D_{t} \cup D_{s}^{*}\right)\right],
\end{aligned}
\end{equation}
where $\mathcal{F}\left(D_{s}^{*} \cup D_{t}\right)$ denotes the model $\mathcal{F}$ trained on the union of $D_{t}$ and $D_{s}^{*}$, the $\mathbb{E}_{D_{t}}$ denotes the risk on the validation set of $D_{t}$.

Ideally, using more data yields better performance. However, the assumption does not always stand in transfer learning scenario due to the domain gap.
The task relationship could be roughly divided into three types: positive correlation, negative correlation, and unrelatedness. If the domain between the source and target tasks are similar, the relationship tends to be positive. If two tasks are dissimilar, the source task may not boost the target task, even hurt the performance. Hence, the essential problem is how to close the gap between two tasks. Some works focus on adversarial training~\cite{Chen2018DomainAF} and data synthesis~\cite{Inoue2018CrossDomainWO}, while we turn to exploit the large-scale public object detection datasets by data selection.

We find that models of GAIA under large-scale pre-training implicitly learn the domain representation in both upstream and downstream datasets, suggesting that it is possible to measure the domain similarity. 
We visualize the output features of {\it person} in {\it fc7} layer from GAIA in Figure~\ref{fig:gaia_datasets_cluster}, and the following characteristics can be observed. 
First, features from common datasets like COCO, Objects365, and Open Images, are apt to occupy diverging space, while features from specific datasets like Caltech, CityPersons, Comic, and Watercolor usually lie in a small and compact space respectively. 
Second, feature spaces of similar datasets are close to each other. For instance, the Comic shares almost the same space with Watercolor. 
Third, there are noticeable overlaps of feature space between common and specific datasets, indicating that common datasets may contain similar data as those in specific datasets. 

With the ability of domain clustering powered by GAIA, we propose to produce the $D_{s}^{*}$ by data selection.
To be specific, for each image in $D_{s} = \{I_{s1}, I_{s2}, \cdots, I_{sP}\}$ and $D_{t} = \{I_{t1}, I_{t2}, \cdots, I_{tQ}\}$ we compute a represent vector for its every category, where $P$ and $Q$ are the number of images in $D_{s}$ and $D_{t}$, respectively. Those categories are defined in the unified label space $\cup L = \{c_{\cup 1}, c_{\cup 2}, \cdots\}$. For each image and each category, the represent vector $V(I_{i}, c_{\cup p})$ could be obtained by averaging the output features in {\it fc7} layer from all instances.
Then we find the most relevant data for each category by using two alternative strategies based on the cosine distances between $V(I_{si}, c_{\cup p})$ and $V(I_{tj}, c_{\cup p})$:
\begin{itemize}
    \item \textbf{top-k}. Choosing top-k images from $D_{s}$ for each $I_{tj}$.
    \item \textbf{most-similar}. Retrieving the most similar images in all $P \times Q$ relation pairs.
\end{itemize}
We collect relevant images until the total number meets the expectation,  for instance,  $\left|D_{s}^{*}\right|=1000$.


\section{Experiments}
\subsection{Datasets}
\label{sec:datasets}

\noindent \textbf{Upstream Datasets.}
GAIA is trained on the union of Open Images~\cite{Kuznetsova2020TheOI}, Objects365~\cite{shao2019objects365}, MS-COCO~\cite{Lin2014MicrosoftCC}, Caltech~\cite{Dollr2012PedestrianDA}, and CityPersons~\cite{Zhang2017CityPersonsAD} under the unified label space. Open Images, Objects365, and MS-COCO are the common detection datasets, containing 500, 365, and 80 categories, respectively. We use the Open Images 2019 challenge split with 1.7 millions images as train set and 40k images as validation. For Objects365, following the official protocol, we use 600k for training and 30k for validation. For COCO, following the standard protocol, 115k subset is used to train, and 5k is used as {\it minival}. Caltech and CityPersons are two specific datasets for pedestrian detection, containing 42k and 3k train set, 4k and 0.5k validation, respectively. These upstream datasets result in a unified label space with 700 categories.

\begin{table}[ht]
	\vspace{-0.1in} 
	\caption{Setting of search space that surrounds different model anchors. Each architecture parameter is sampled from minimum to maximum with certain step. $W$[0] denotes the width of the network stem.}
	\centering 
	\footnotesize
	\begin{tabular}{l|c|c|c}
	 & AR50  & AR77 & AR101 \\
    \hline
    \hline
    $D_{min}$ & [2,2,4,2] & [2,2,11,2] & [2,2,17,2] \\
    $D_{anchor}$ & [3,4,6,3] & [3,4,15,3] & [3,4,23,3] \\
    $D_{max}$ & [4,6,8,4] & [4,6,19,4] & [4,6,29,4] \\
    $D_{step}$ & [1,2,2,1] & [1,2,4,1] & [1,2,6,1] \\
    \hline
    $W_{min}$ &  \multicolumn{3}{c}{[32,48,96,192,384]}  \\
    $W_{anchor}$ & \multicolumn{3}{c}{[64,64,128,256,512]}  \\
    $W_{max}$ & \multicolumn{3}{c}{[64,80,160,320,640]}  \\
    $W_{step}$ & \multicolumn{3}{c}{[16,16,32,64,128]}  \\
    \hline
    $S_{min}$ & 400  & 480 & 560 \\
    $S_{anchor}$ & 560 & 640 & 720  \\
    $S_{max}$ & 720 & 800 & 880 \\
    $S_{step}$ & 80  & 80 & 80 \\
    \hline
        
	\end{tabular}
	\label{tab:methods:anchor_space}
	\vspace{-0.1in} 

\end{table}

\begin{table*}[t]
	\caption{Results of models with different FLOPs on COCO {\em minival}. 
	All models are trained with $1\times$ scheduler if not specified. Models with ``*" are trained with $3\times$ scheduler. The FLOPs is calculated with input size of [3,Scale,Scale], and the latency is measured on the entire validation set of COCO on V100 with batchsize of 1.}
	\centering 
	\footnotesize
	\setlength\tabcolsep{5.5pt}
	\begin{tabular}{c|c|c|c|c|c|c|ccc|ccc}
		Group & Pre-train & Scale & Depth & Width & FLOPs & Latency & AP & AP$_{50}$ & AP$_{75}$ & AP$_S$ & AP$_M$ & AP$_L$  \\
		\hline
		\hline
		ResNet50 & \multirow{2}{*}{ImageNet}& 800 & [3,4,6,3] & [64, 64, 128, 256, 512] & 137.4B & 39ms & 37.08 & 59.52 & 39.74 & 22.63 & 40.92 & 46.51 \\
		ResNet101 & & 800 & [3,4,23,3] & [64, 64, 128, 256, 512] & 188.5B & 51ms & 39.41 & 61.62 & 43.04 & 24.42 & 43.61 & 50.53 \\
		\hline
		ResNet50$*$ & \multirow{2}{*}{ImageNet}& 800 & [3,4,6,3] & [64, 64, 128, 256, 512] & 137.4B & 39ms & 39.78 & 60.47 & 42.23 & 23.27 & 42.57 & 50.04 \\
		ResNet101$*$ & & 800 & [3,4,23,3] & [64, 64, 128, 256, 512] & 188.5B & 51ms & 41.75 & 62.24 & 45.77 & 25.28 & 45.61 & 54.39 \\
		\hline
		ResNet50 & \multirow{2}{*}{GAIA}& 800 & [3,4,6,3] & [64, 64, 128, 256, 512] & 137.4B & 39ms & 42.91 & 64.43 & 47.12 & 28.01 & 47.07 & 53.81 \\
		ResNet101 & & 800 & [3,4,23,3] & [64, 64, 128, 256, 512] & 188.5B & 51ms & 46.07 & 67.28 & 50.64 & 29.31 & 50.45 & 58.50 \\
		\hline
		30-45B & \multirow{10}{*}{GAIA}& 400 & [4,4,8,4] & [48,48,96,192,384] & 44.3B & 17ms & 38.21 & 58.73 &  40.62 & 18.15 & 42.18 & 54.79 \\
		45-60B & & 480 & [4,6,8,4] & [48,48,96,256,384] & 59.4B & 19ms & 40.44 & 61.07 &  43.77 & 21.65 & 43.66 & 55.96 \\
		60-75B  & & 560 & [4,6,15,4] & [48,80,96,192,512] & 74.4B & 26ms & 42.59 & 64.01 & 46.07 & 24.82 & 46.78 & 57.42 \\
		75-90B  & &560 & [4,6,21,4] & [64,80,96,192,512] & 88.1B & 28ms & 43.60 & 64.81 & 47.34 & 24.97 & 47.85 & 58.62 \\
        90-105B  & &560 & [4,6,21,4] & [64,80,160,192,512] & 101.1B & 30ms& 44.05 & 65.15 & 47.91 & 25.55 & 48.69 & 59.09 \\
        105-120B  & &640 & [4,6,21,4] & [64,80,160,192,512] & 119.2B & 33ms & 44.46 & 65.71 & 47.49 & 26.77 & 48.31 & 57.21 \\
        120-135B  & &720 & [3,4,23,3] & [64,64,128,192,640] & 133.9B & 38ms & 45.27 & 66.64 & 49.47 & 27.97 & 49.44 & 57.89 \\
        135-150B  & &800 & [4,6,23,3] & [48,48,96,192,640] & 149.1B & 44ms & 45.55 & 66.89 & 50.08 & 28.96 & 49.94 & 58.01 \\
        150-180B  & &800 & [3,4,23,3] & [64,64,96,256,512] & 178.7B & 47ms & 46.02 & 67.40 & 50.52 & 28.80 & 50.37 & 59.01 \\
        180-210B  & &880 & [3,4,25,4] & [48,48,96,256,384] & 209.8B &53ms & 46.41 & 67.98 & 50.88 & 29.83 & 50.83 & 58.32 \\ 
        \hline
        
	\end{tabular}\vspace{0.1cm}
	\label{tab:exp:main_results}
	\vspace{-0.1in}

\end{table*}

\noindent \textbf{Downstream Tasks.}
We conduct extensive experiments on 
Universal Object Detection Benchmark (UODB)~\cite{Wang2019TowardsUO}. Note that UODB is not used in upstream training phase and consists of 10 diverse sub-datasets: Pascal VOC~\cite{Everingham2009ThePV}, WiderFace~\cite{Yang2016WIDERFA}, KITTI~\cite{Geiger2012AreWR}, LISA~\cite{Mgelmose2012VisionBasedTS}, DOTA~\cite{Xia2018DOTAAL}, Watercolor~\cite{Inoue2018CrossDomainWO}, Clipart~\cite{Inoue2018CrossDomainWO}, Comic~\cite{Inoue2018CrossDomainWO}, Kitchen~\cite{Georgakis2016MultiviewRD} and DeepLesions~\cite{Yan2018DeepLG}. We follow the official data split and metric for all above datasets. Details of each dataset are documented in the Supplementary Material section A.

\subsection{Implementation Details}

\subsubsection{Architecture Space}
In all experiments, we use the Faster RCNN~\cite{Ren2015FasterRT} with FPN~\cite{Lin2017FeaturePN} as the base framework. ResNet~\cite{He2016DeepRL} is adopted as the network prototype in GAIA, and both the task-agnostic unification and task-specific adaptation are applied on it. 
The reason for choosing ResNet is that ResNet is still the most paradigmatic network architecture in object detection and is especially popular in real-world applications.
For {\it anchor-based progressive shrinking} (ABPS) during training stage, we choose 3 model anchors in our experiments for simplicity and the corresponding search spaces are shown in Table~\ref{tab:methods:anchor_space}. We refer to these model anchors as AR50, AR77 and AR101. In each iteration during training, we randomly pick an available model that obeys one of the prescribed rules in search space for optimization. The rules includes: models with the maximum depth, models with the minimum width and other likenesses.
Details of more prescribed rules are shown in the Supplementary Material section B.

\subsubsection{Upstream Training}
The training of the supernet starts with AR101 for 24 epochs, then the search space shrinks to AR77 and finally to AR50, and we finetune each search space for 13 epochs. Each time we shrink search space, we add a single epoch of warm-up which is essential.
We follow the training setting defined in Detectron2~\cite{wu2019detectron2} except the input scales are adjusted to $[S_{min}:S_{step}:S_{max}]$ for each sub search space. By convention, no other data augmentation is applied except the standard horizontal flipping. The initial learning rate is set 0.00125 per image and is decayed by a factor of 10 at 16 and 21 epoch. For finetuning in the shrunk search space, the learning rate restarts with the initial value and is decayed by a factor of 10 at 8 and 11 epoch. The supernet of GAIA is trained from scratch, thus we apply the sync-BN which is essential as proved in~\cite{He2019RethinkingIP}. We use SGD to optimize the training loss with 0.9 momentum and 0.0001 weight decay.
IoU-sampling~\cite{pang2019libra} is applied to make sure that subnets could learn balanced knowledge across object scales. 
Since the supernet of GAIA are trained to learn a great number of categories, we find that the gradients derived from class-specific supervision are diluted over time compared to that from class-agnostic supervision. To alleviate the issue, we multiply the loss weights of head by a factor of 5.


\subsubsection{Downstream Fine-Tuning}
Given a downstream dataset and categories of interests, we apply the TSAS to find the most appropriate architecture and extract corresponding weights from supernet.
Besides, we conduct a weight surgery on the weights of the last {\it fc} layer in head to focus on related categories. For categories included in the unified label space of GAIA, we keep the pre-trained weights. For categories that are not in the label space, we find their closest neighbors based on the word vectors, and take the weights for initialization.

The initial learning rate of downstream finetuning is set 0.0001875 per image and is decayed by a factor of 5 after 8 and 11 epoch. We train for 13 epochs in total and all the other configuration is consistent with the training setting of supernet.
As for the fast-finetuning in TSAS, we use a warm-up for one epoch and train for 2 epochs. The initial learning rate of downstream finetuning is set 0.0001875 per image and is decayed by a factor of 10 after the first epoch.

\begin{table*}[ht]
	\caption{Results of models with different whistles and bells on COCO {\em minival}.
	All models are trained with $1\times$ scheduler. ``ImageNet'' and ``GAIA'' denote the   pre-training datasets respectively. The architecture of GAIA-TSAS in table is \{$D$: [4,6,23,3], $W$:[48,48,128,192,384], $S$: 800\}.}
	\centering 
	\footnotesize
	\begin{tabular}{c|cccc|ccc|ccc|c}
		Backbone & ImageNet & GAIA & DCN & Cascade-Head  & AP & AP$_{50}$ & AP$_{75}$ & AP$_S$ & AP$_M$ & AP$_L$ & Latency \\
		\hline
		\hline
		\multirow{8}{*}{ResNet50}  & \checkmark & & & & 37.0 & 59.5 & 39.7 & 22.6 & 40.9 & 46.5 & \multirow{2}{*}{39ms}\\
		 &  & \checkmark & &   & 42.9 & 64.4 & 47.1 & 28.0 & 47.1 & 53.8 & \\
         \cline{2-12}
         & \checkmark & & \checkmark &  & 40.8 & 63.7 & 44.3 & 25.6 & 44.4 & 52.3 & \multirow{2}{*}{44ms}\\
         &  & \checkmark & \checkmark &  & 44.8 & 66.4 & 49.2 & 29.0 & 48.6 & 57.3 & \\
         \cline{2-12}
          &  \checkmark & & & \checkmark & 40.9 & 59.7 & 44.4 & 24.1 & 44.5 & 52.3 & \multirow{2}{*}{47ms}\\
          &  & \checkmark & & \checkmark   & 45.8 & 64.6 & 50.1 & 29.7 & 49.6 & 58.4 & \\
          \cline{2-12}
          &  \checkmark &  & \checkmark & \checkmark   & 44.9 & 64.8 & 49.2 & 27.2 & 49.2 & 59.4 & \multirow{2}{*}{53ms}\\
          &  & \checkmark & \checkmark & \checkmark & 47.9 & 66.9 & 52.6 & 31.7 & 51.5 & 61.9 & \\
          \hline
          GAIA-TSAS & & \checkmark & \checkmark & \checkmark & 49.1 & 68.0 & 54.0 & 30.5 & 53.4 & 65.0 & 55ms\\
		\hline

	\end{tabular}\vspace{0.1cm}
	\label{tab:exp:whistles_and_bells}
	\vspace{-0.1in}

\end{table*}

\begin{table}[th]
	\caption{Results of GAIA on other upstream datasets using ResNet50. We follow the official protocol, \ie, mmAP for Objects365, mAP for Open Images, and MR$^{-2}$(Miss Rate, lower is better) for Caltech and CityPersons. ``TSAS'' denotes that an architecture with similar latency as ResNet50 is applied. }
	\centering 
	\footnotesize
	\begin{tabular}{c|cc|c|c}
	\multirow{2}{*}{Dataset}&\multicolumn{2}{c|}{Pre-train}&\multirow{2}{*}{TSAS}&\multirow{2}{*}{Metric}  \\
	                         &   ImageNet & GAIA             &  &  \\
    \hline
    \hline
    \multirow{3}{*}{Objects365}& \checkmark &           & & 21.5 \\
                               &            &\checkmark& & 24.0 \\
                               &            &\checkmark &\checkmark & \textbf{26.1} \\
    \hline
    \multirow{3}{*}{Open Images}& \checkmark &           & & 52.2 \\
                               &            &\checkmark& & 59.5 \\
                               &            &\checkmark &\checkmark & \textbf{62.4} \\
    \hline
    \multirow{3}{*}{Caltech}& \checkmark &           & & 5.5 \\
                            &            &\checkmark& & 2.2 \\
                            &            &\checkmark &\checkmark & \textbf{1.7} \\
    \hline
    \multirow{3}{*}{CityPersons}& \checkmark &           & & 14.7 \\
                                &            &\checkmark& & 11.1 \\
                                &            &\checkmark &\checkmark & \textbf{10.4} \\
    \hline
	\end{tabular}
	\label{tab:exp:other_dataset}
\end{table}

\begin{table*}[ht]
	\caption{Results of GAIA on UOBD datasets. ${\dagger}$ stands for our re-implemented baseline. ``COCO'' and ``GAIA'' denote the pre-training datasets, respectively.}
	\centering 
	\footnotesize
	\begin{tabular}{c|c|c|c|c|c|c|c|c|c|c|c}
	Method & KITTI & VOC & WiderFace & LISA & Kitchen & DOTA & DeepLesion & Comic & Clipart & Watercolor & Avg. \\
    \hline
    \hline
    Baseline~\cite{Wang2019TowardsUO}& 64.3 & 78.5& 48.8      & 88.3 & 87.7    & 57.5 & 51.2       & 45.8  & 32.1    & 52.6       & 60.7 \\
    \hline
    DA~\cite{Wang2019TowardsUO}& 68.0  & 82.4& 51.3      & 87.6 & 90.0    & 56.3 & 53.4       & 53.4  & 55.8    & 60.6       & 65.9 \\
    \hline
    \hline
Baseline$^{\dagger}$&67.1& 81.5& 62.1      & 90.0 & 89.5    & 68.3 & 57.4       & 45.5  & 31.2    & 53.4       & 64.6 \\
    \hline
    COCO   & 70.2 & 84.2 & 61.4      & 88.6 & 88.7    & 64.7 & 52.4       & 51.6  & 56.9    & 56.1       & 67.5 \\
    \hline
    GAIA   & 72.9 & 85.9 & 62.6      & 91.2 & 89.8    & 69.2 & 59.4       & 57.0  & 67.9    & 63.5       & 71.9 \\
    \hline
    GAIA-TSAS&\textbf{75.6}& \textbf{87.4} & \textbf{62.7}      & \textbf{92.1} & \textbf{90.1}    & \textbf{70.8} & \textbf{62.1}       & \textbf{61.1}  & \textbf{72.2}    & \textbf{69.7}       & \textbf{74.4} \\
    \hline
	\end{tabular}
	\label{tab:exp:uodb_dataset}
\end{table*}

\begin{table}[th]
	\caption{Results of task-specific data selection (TSDS). Three strategies, including random, top-k, and most-similar, are used.}
	\centering 
	\footnotesize
	\setlength\tabcolsep{5pt}
	\begin{tabular}{c|c|c|c|c|c}
	pre-train & TSDS & KITTI & Comic & Watercolor & Average \\
    \hline
    \hline
    COCO    &-                   & 44.9 & 36.4  & 45.3       & 42.2 \\
    \hline
    \multirow{4}{*}{GAIA}&-      & 45.4 & 46.9  & 51.1       & 47.8 \\
    \cline{2-6}
                        &random  & 40.3 & 46.1  & 49.9       & 45.4 \\
    \cline{2-6}
                        &top-k   & 46.7 & 48.0  & 51.2       & 48.6 \\
    \cline{2-6}
                    &most-similar& \textbf{48.6} & \textbf{48.2}  & \textbf{53.6}      & \textbf{50.1} \\
    \hline
	\end{tabular}
	\label{tab:exp:few_datapoints}
\end{table}

\subsection{Results on COCO Dataset}
Taking COCO dataset as an example, we demonstrate how GAIA is capable of generating high-quality models powered by data unification and architecture adaptation. First, we compare the results of the vanilla ResNet50 and ResNet101 trained with different weight initialization. As shown in Table~\ref{tab:exp:main_results}, models with GAIA pre-training yield huge improvements over models with ImageNet pre-training, which is $5.83\%$ for ResNet50 and $6.66\%$ for ResNet101. 
Since the data of COCO dataset are included in the data pool of supernet, we also compare the results of GAIA with models trained for $3\times$ with ImageNet pre-training for fairness. The improvements are still considerable($+3.23\%$ and $+4.22\%$), indicating that data from other sources are of great help. With the help of TSAS, the improvements are further boosted to $5.49\%$ and $4.66\%$. 

In addition, GAIA is able to produce models across a wide latency range efficiently. Since there are no pre-trained weights for models other than ResNet50 and ResNet101, models with customized architecture are obliged to be trained from scratch. Within $1\times$ the training time, models trained from GAIA outperform models trained from scratch $12.67\%$ on average as shown in Figure~\ref{fig:intro:main_results_latency}.

We also conduct experiments to see whether models generated from GAIA are compatible with whistles and bells. We select DCN~\cite{Dai2017DeformableCN} and Cascade-RCNN~\cite{Cai2018CascadeRD} which are two of the most effective methods in object detection. As displayed in Table~\ref{tab:exp:whistles_and_bells}, the gain from GAIA is congruent with these methods. With the help of TSAS, we achieve an AP of $49.1\%$ with only $1\times$ scheduler while keeping the latency almost unchanged.

\subsection{Results on Other Upstream Datasets}
We also provide the results in other seen datasets in Table~\ref{tab:exp:other_dataset}. Objects365, Open Images, Caltech, and CityPersons are parts of the pre-training datasets for GAIA. We build their baseline from widely-used ImageNet pre-training, and train them individually until they converge and the performances stop growing. Then we apply downstream finetuning for all datasets from GAIA pre-training. We find that GAIA outperforms the baseline in these four datasets by $2.5\%$, $8.8\%$, $3.3\%$, and $3.6\%$, respectively. The significant improvement in Open Images is mainly caused by the gain from those long-tail categories. Moreover, with the ability of task-specific architecture selection (TSAS), GAIA yields additional $0.5\% \sim 2.9\%$ improvements.
Detailed comparisons with state-of-the-art are available in the supplementary material.

\subsection{Transfer Learning on Downstream Datasets}
To evaluate the generalizability of GAIA, we conduct experiments on downstream datasets from  UODB~\cite{Wang2019TowardsUO}. Table~\ref{tab:exp:uodb_dataset} reports the mAP and GAIA shows great success on various datasets. It can be seen that our re-implemented ImageNet pre-trained baseline achieves $64.6\%$ mAP on average which is $3.9\%$ higher than the UODB baseline, and the COCO pre-training brings $2.9\%$ improvements. Therefore, these counterparts are strong enough to validate the effectiveness of GAIA. Empowered by the unified label space and large-scale dataset pre-training, GAIA can steadily improve the performance by $4.4\%$ on average. Furthermore, TSAS of GAIA yields another $2.5\%$ improvements overall.

\subsection{Data Selection for Data-scarce Scenarios}
In this section, we evaluate our proposed task-specific data selection (TSDS) in few datapoints settings. From the unseen UODB data source, we pick three datasets with small label space for convenience. We start our investigation with 10 images per datasets as the baseline. The images are randomly sampled from the corresponding dataset while making each category have at least 2 images.
As shown in Table~\ref{tab:exp:few_datapoints}, the average mAP of GAIA without TSDS outperforms the COCO pre-trained baseline by $5.6\%$. With data selection strategies, it is obvious that choosing relevant data is crucial. Randomly selecting data is harmful to the performance because it introduces much out-domain data and disturbs the target domain learning. It is also nice to see that top-k and most-similar strategies bring additional gains by $0.8\% \sim 2.3\%$ on average, showing the benefit of large-scale pre-trained GAIA for the data selection.

\section{Conclusion}
In this work, we revisit the efficacious generalizability of transfer learning and its limitation to downstream customization, and harmonize the gap between generalist and specialist models.
We present a transfer learning system named GAIA, which could automatically and efficiently give birth to specialized solutions according to heterogeneous downstream needs.
Constructing GAIA involves a heavy workload, thus it leaves much space for future work to make it better. It could also be extended to more architectures like MobileNetV3, more frameworks like GAIA-YOLO, and more vision tasks like GAIA-Seg. 
In the end, we sincerely hope that our work could substantially help more practitioners in the community of object detection. 

\section{Acknowledgements}
This work was supported in part by the Major Project for New Generation of AI (No.2018AAA0100400), the National Natural Science Foundation of China (No. 61836014, No. 61773375).

{\small
\bibliography{egbib}
\bibliographystyle{ieee_fullname}
}


\appendixpage
\appendix
\section{Details of datasets}
We list the information of each dataset in Table~\ref{tab:exp:dataset}, including number of categories, data scale of training and validation sets, and metrics. 

\begin{table}[ht]
\caption{Details about the different datasets. mAP, mmAP and MR$^{-2}$ are abbreviations of the mean Average Precision at overlap 0.5, mean mAP over overlap ranging in [0.5, 0.95], and log average Miss Rate over false positives per image ranging in [10$^{-2}$, 10$^{0}$].}
	\centering 
	\footnotesize
	\begin{tabular}{c|c|c|c|c}
	Dataset & Category & Train & Validation & Metric \\
    \hline
    \hline
    Open Images& 500    & 1.7m  & 40k        & mAP    \\
    \hline
    Objects365& 365    & 600k  & 30k        & mmAP    \\
    \hline
    COCO      &80    & 115k  & 5k        & mmAP    \\
    \hline
    Caltech   &1    & 42k  & 3k        & MR$^{-2}$    \\
    \hline
    CityPersons&1    & 3k  & 0.5k        & MR$^{-2}$    \\
    \hline
    VOC& 20    & 16k  & 5k        & mAP    \\
    \hline
    WiderFace& 1    & 13k  & 3k        & mAP    \\
    \hline
    KITTI& 3    & 4k  & 4k        & mAP    \\
    \hline
    LISA& 4    & 8k  & 2k        & mAP    \\
    \hline
    DOTA& 15    & 14k  & 5k        & mAP    \\
    \hline
    Watercolor& 6    & 1k  & 1k        & mAP    \\
    \hline
    Clipart& 20    & 0.5k  & 0.5k        & mAP    \\
    \hline
    Comic& 6    & 1k  & 1k        & mAP    \\
    \hline
    Kitchen& 11    & 5k  & 2k        & mAP    \\
    \hline
    DeepLesions& 1    & 28k  & 5k        & mAP    \\
    \hline
	\end{tabular}
	\label{tab:exp:dataset}
\end{table}

\section{Rules for architecture selection}
As mentioned in the Section4.2.1 of paper, there are some prescribed rules for architecture selection. We denote the minimum total depth and maximum total depth as $d_{min}$ and $d_{max}$. The depth of model is denoted as $d$ and we have $d'=d_{max} - d_{min}$ for simplicity. The pool of rules and respective sampling probability are shown as follow:
\begin{itemize}
    \item models with $d=d_{min}$, $p=\frac{1}{8}$.
    \item models of the $d=d_{min}+0.25d'$, $p=\frac{1}{8}$.
    \item models of the $d=d_{min}+0.5d'$, $p=\frac{1}{8}$.
    \item models of the $d=d_{min}+0.75d'$, $p=\frac{1}{8}$.
    \item models of the $d=d_{max}$, $p=\frac{1}{8}$.
    \item random models, $p=\frac{3}{8}$.
\end{itemize}

\section{Visualization results on each dataset}
\begin{figure*}[t]
\centering
\begin{subfigure}[b]{0.95\linewidth}
\centering
\includegraphics[width=\linewidth]{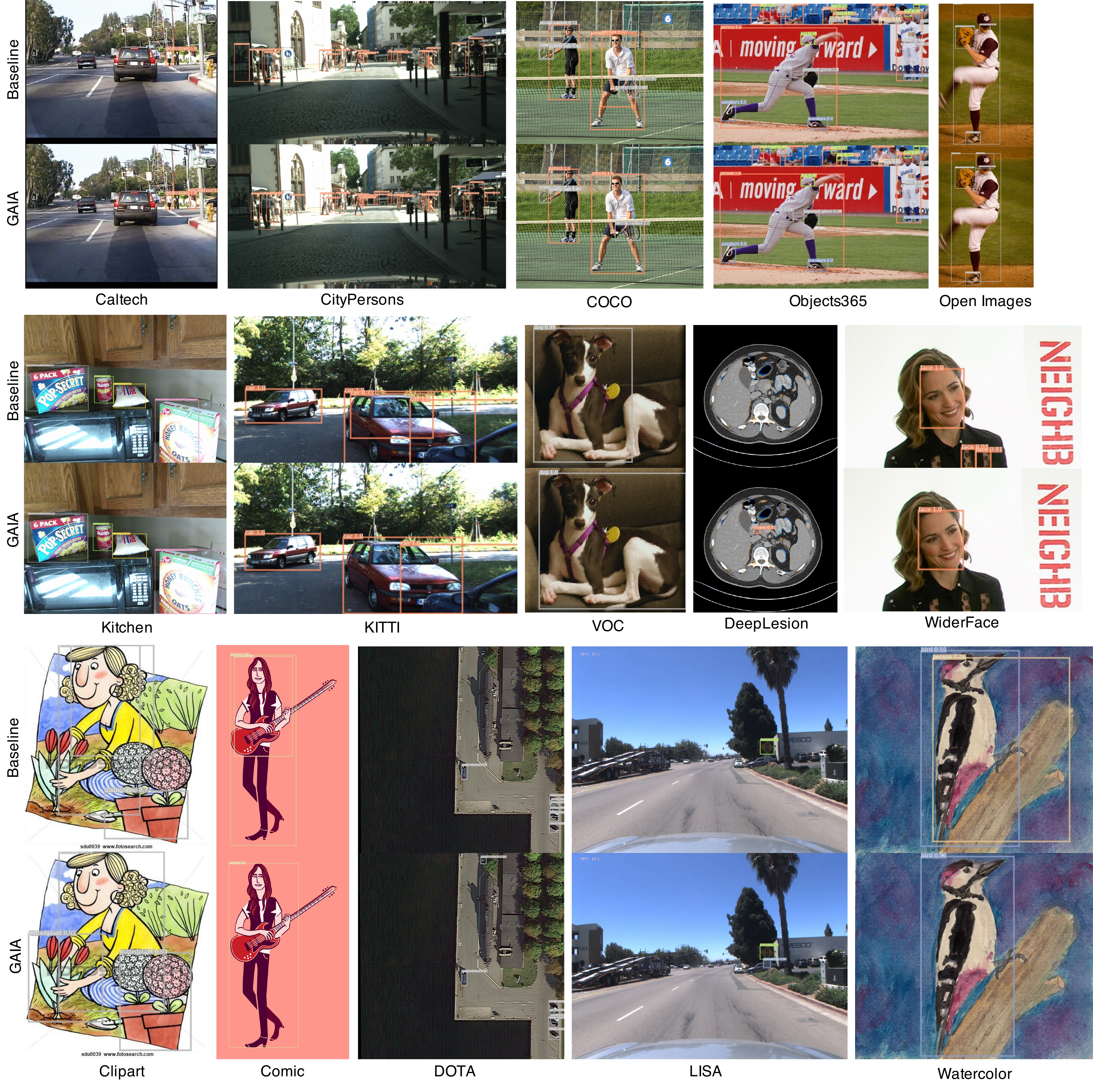}
\end{subfigure}
\caption{
Examples of detection results from ImageNet baseline and GAIA on each dataset. 
}
\label{fig:detections_15in1}
\end{figure*}
We visualize the detection results on each dataset, as demonstrated in Figure~\ref{fig:detections_15in1}.

\section{The adapted architectures to each dataset}
We list the selected architecture for each downstream task(Table 5 in paper) in Table~\ref{tab:arch}. ``$\dagger$'': In CityPersons dataset, the default input size is $1024\times2048$, thus we build our search space of input scale surrounding the default input size with a step of 128 pixels.
\begin{table}[ht]
\caption{Details about the adapted architectures.}
    \centering 
    \footnotesize
    \begin{tabular}{c|c|c|c}
    Dataset & Scale & Depth & Width \\
    \hline
    \hline
    Open Images&640   &[4,6,29,4]  &[64,80,160,192,640] \\
    \hline
    Objects365&720    &[3,4,23,3]  &[64,64,128,192,640]\\
    \hline
    COCO      &720   & [3,4,23,3]  &[64,64,128,192,640]\\
    \hline
    Caltech   &880    &[2,4,17,2]  &[48,48,128,256,640] \\
    \hline
    CityPersons$^{\dagger}$& 1152  &[3,2,4,3]  &[64,64,96,192,384] \\
    \hline
    VOC       &640    &[3,4,29,4]  &[64,64,128,256,512] \\
    \hline
    WiderFace &880    &[4,4,4,2]  &[64,64,96,192,384] \\
    \hline
    KITTI     &880    &[3,4,6,3]  &[48,64,96,192,384] \\
    \hline
    LISA      &720    &[2,4,17,3]  &[64,64,128,192,512] \\
    \hline
    DOTA      &880    &[4,6,4,2]  &[32,48,96,192,512]\\
    \hline
    Watercolor&640    &[3,2,29,3] &[48,80,96,256,640]\\
    \hline
    Clipart   &640    &[3,6,17,3] &[32,64,128,320,640]\\
    \hline
    Comic    &640    &[2,6,17,2] &[48,64,160,320,512]\\
    \hline
    Kitchen  &720    &[3,6,23,2] &[48,48,160,192,512]\\
    \hline
    DeepLesions&720    &[4,4,17,2] &[32,80,96,256,512]\\
    \hline
    \end{tabular}
    \label{tab:arch}
\end{table}

\section{The selected data for each downstream task}
\begin{figure*}[t]
\centering
\begin{subfigure}[b]{0.8\linewidth}
\centering
\includegraphics[width=\textwidth]{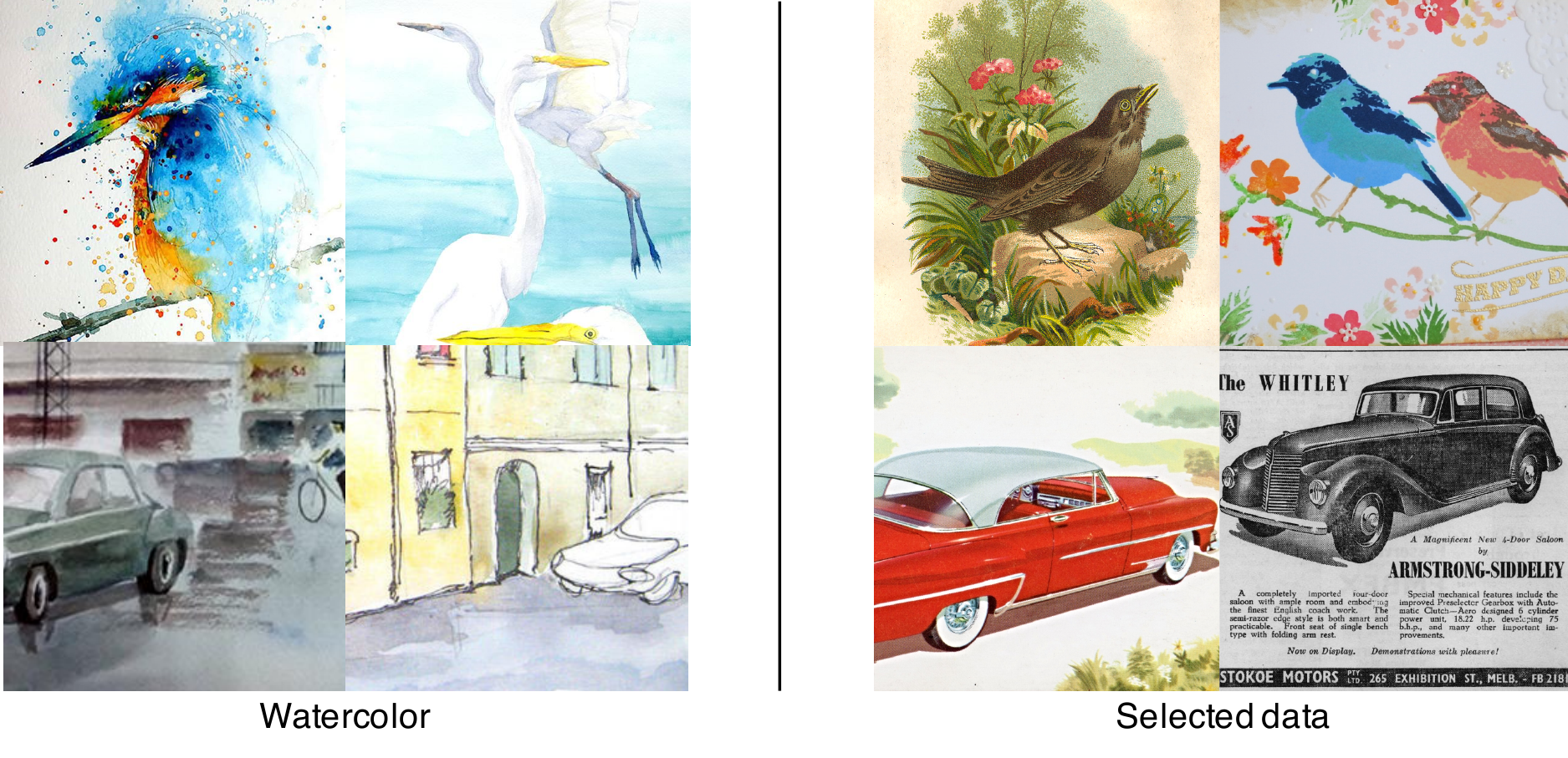}
\label{fig:ds_watercolor}
\end{subfigure}
\begin{subfigure}[b]{0.8\linewidth}
\centering
\includegraphics[width=\textwidth]{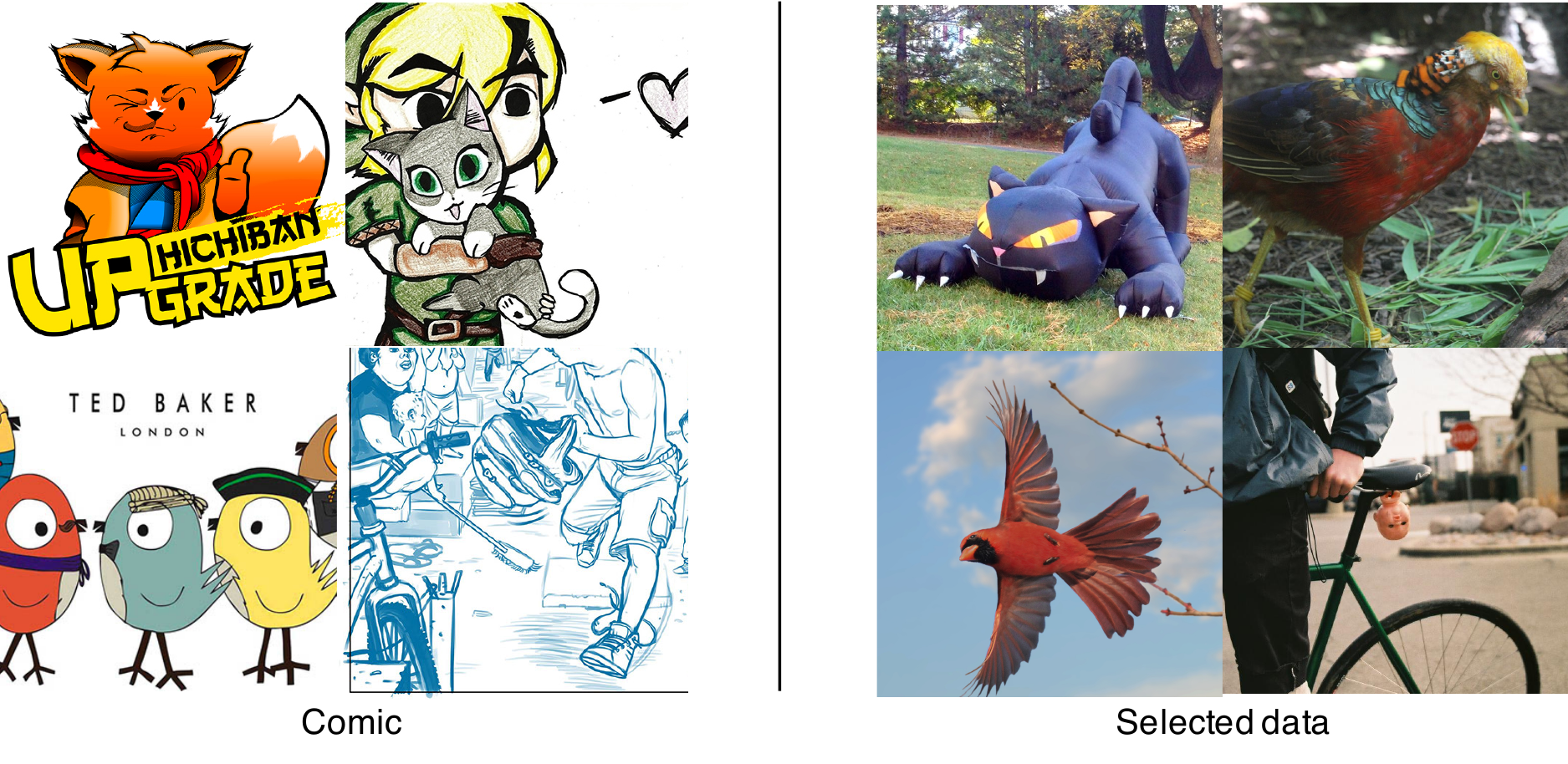}
\label{fig:ds_comic}
\end{subfigure}
\begin{subfigure}[b]{0.8\linewidth}
\centering
\includegraphics[width=\linewidth]{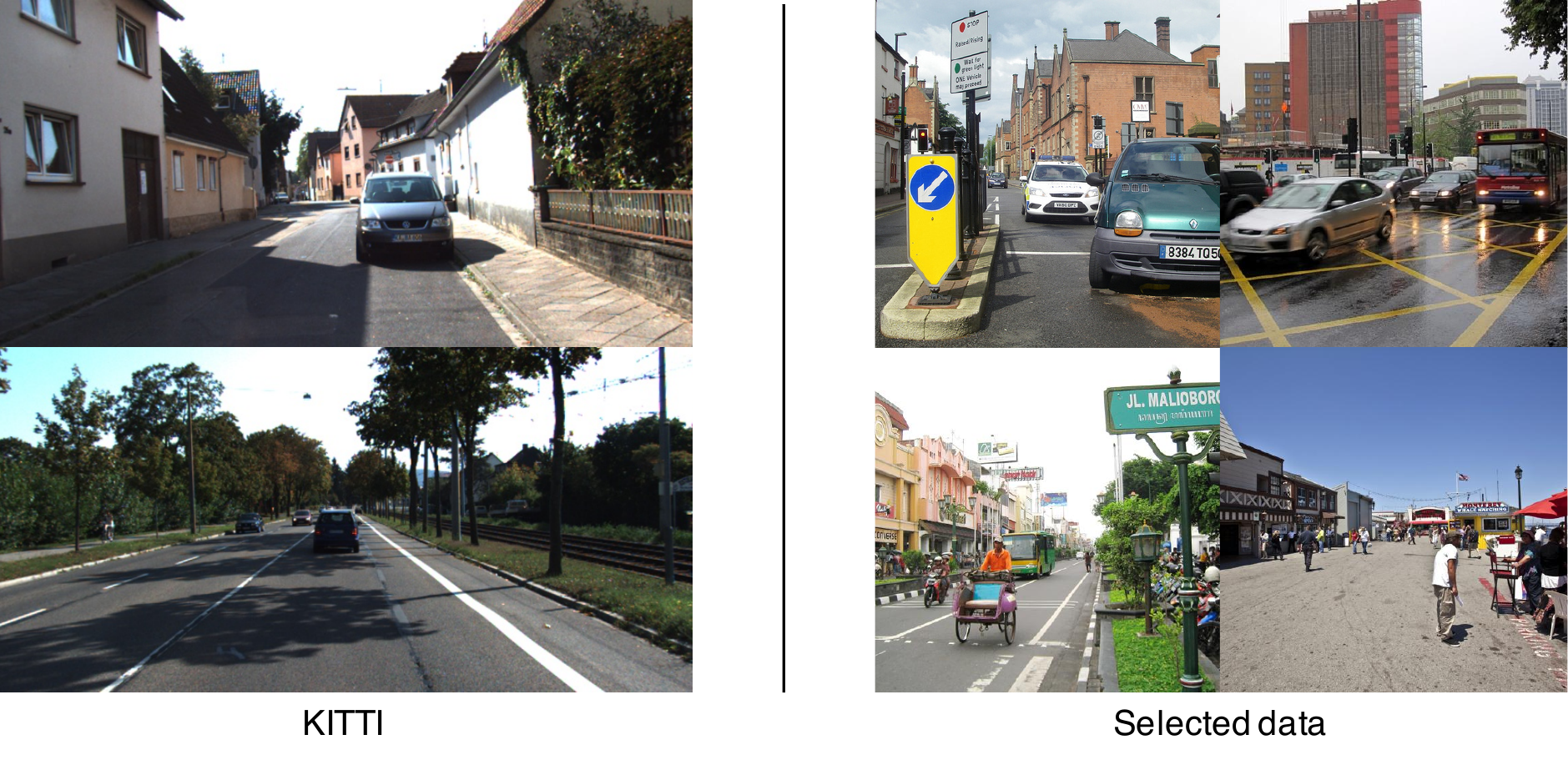}
\label{fig:ds_kitti}
\end{subfigure}
\caption{
Examples of data selection results. From top to bottom: Watercolor, Comic, and KITTI. From left to right: downstream tasks data and their corresponding images selected from upstream datasets. 
}
\label{fig:ds}
\end{figure*}

Given few images from downstream tasks as query, we show the relevant data collected by GAIA in Figure~\ref{fig:ds}. (Table 6 in paper)

\end{document}